\documentclass[11pt,letterpaper]{article}
\usepackage{naaclhlt2015}
\usepackage{times}
\usepackage{url}
\usepackage{latexsym}
\usepackage{comment}
\usepackage{amsmath,amsthm,amssymb}
\usepackage{graphicx}
\usepackage{multirow}
\usepackage{color}

\title{Retrofitting Word Vectors to Semantic Lexicons}
\author{Manaal Faruqui  \quad Jesse Dodge \quad Sujay K. Jauhar \\
  \textbf{Chris Dyer \quad Eduard Hovy \quad Noah A. Smith}\\
  Language Technologies Institute \\
  Carnegie Mellon University \\
  Pittsburgh, PA, 15213, USA \\
{\tt \{mfaruqui,jessed,sjauhar,cdyer,ehovy,nasmith\}@cs.cmu.edu}
}

\begin{document}
\maketitle
\begin{abstract}
  Vector space word representations are learned from distributional information of words
  in large corpora. Although such statistics are semantically informative, they
  disregard the valuable information that is contained in semantic lexicons such as WordNet, FrameNet,
  and the Paraphrase Database.
  This paper proposes a method for refining vector space representations using relational information from semantic lexicons by encouraging linked words to have similar vector representations, and it makes no assumptions about how the input vectors were constructed.
   Evaluated on a battery of standard lexical semantic evaluation tasks in several languages, we obtain substantial improvements starting with a variety of word vector models. Our refinement method outperforms prior techniques for incorporating semantic lexicons into word vector training algorithms.
   \end{abstract}

\section{Introduction}
\label{sec:intro}

Data-driven learning of word vectors that capture lexico-semantic
information is a technique of central importance in NLP.
These word vectors can in turn be used for identifying semantically related word
pairs~\cite{Turney:2006:SSR:1174520.1174523,Agirre:2009:SSR:1620754.1620758} or as features
in downstream text processing applications \cite{turian:2010,guo2014revisiting}.
A variety of approaches for constructing vector space embeddings of vocabularies are in use, notably including taking low rank approximations of cooccurrence
statistics
\cite{deerwester-90} and using internal representations from neural
network models of word sequences~\cite{Collobert:2008:UAN:1390156.1390177}.

Because of their value as lexical semantic representations, there has been much research on improving the quality of vectors.
\emph{Semantic lexicons}, which provide type-level information about the semantics of words, typically by identifying \textit{synonymy}, \textit{hypernymy},
\textit{hyponymy}, and \textit{paraphrase} relations should be a valuable resource
for improving the quality of word vectors that are  trained solely on unlabeled corpora.
Examples of such resources include WordNet~\cite{miller:1995},
FrameNet~\cite{Baker:1998:BFP:980845.980860} and the Paraphrase
Database~\cite{ganitkevitch2013ppdb}.

Recent work has shown that
by either changing the objective of the word vector training algorithm in neural language models~\cite{Yu:2014,xu2014rc,yan-ecml14,fried2014incorporating} or by relation-specific augmentation of the cooccurence
matrix in spectral word vector models to incorporate semantic knowledge
\cite{Yih:2012,chang-yih-meek:2013:EMNLP},
the quality of word vectors can be improved. However, these methods are limited to particular methods for constructing vectors.

The contribution of this paper is a graph-based learning technique for using
lexical relational resources to obtain higher quality semantic vectors, which we call ``retrofitting.''
In contrast to previous work, retrofitting is applied as a \emph{post-processing step} by running belief propagation on a graph constructed from lexicon-derived relational information to update word vectors (\S\ref{sec:framework}).
This allows retrofitting to be used on pre-trained word vectors obtained using
\textit{any} vector training model.
Intuitively, our method encourages the new vectors to be (i) similar to the
vectors of related word types and (ii) similar to their purely distributional
representations.
The retrofitting process is fast, taking about 5 seconds for a graph of 100,000
words and vector length 300, and its runtime is independent of the original
word vector training model.

Experimentally, we show that our method works well with
different state-of-the-art word vector models,
using different kinds of semantic lexicons
and gives substantial improvements on a variety of benchmarks, while beating the current state-of-the-art approaches for
incorporating semantic information in vector training and
trivially extends to multiple languages.
We show that retrofitting gives consistent improvement in performance on
evaluation benchmarks with different word vector lengths
and show a qualitative visualization of the effect of retrofitting on word
vector quality. The retrofitting tool is available at:
\url{https://github.com/mfaruqui/retrofitting}. 

\section{Retrofitting with Semantic Lexicons}
\label{sec:framework}

\begin{figure}[tb]
  \centering
  \includegraphics[width=\columnwidth]{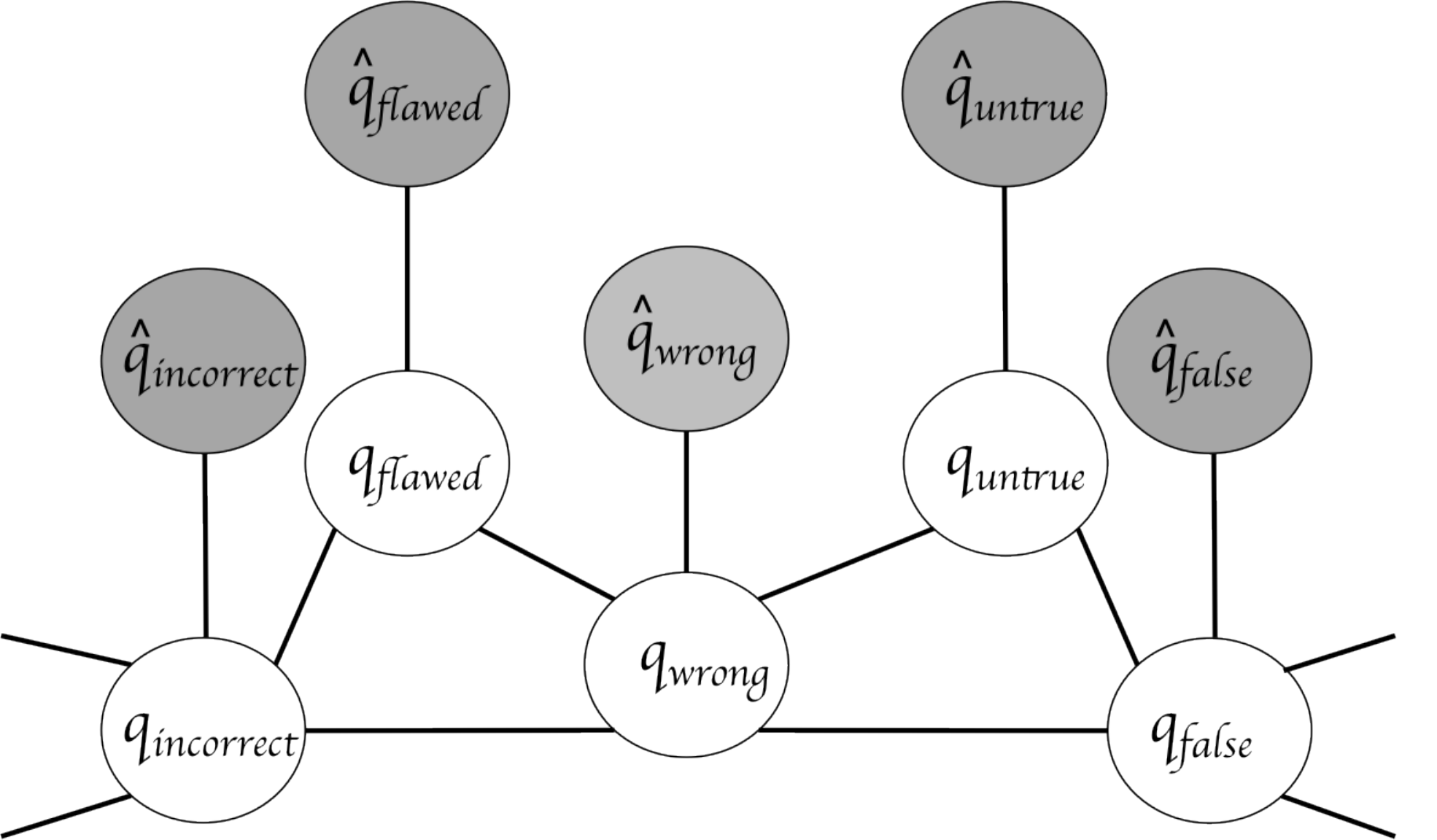}
  \caption{Word graph with edges between related words showing the
    observed (grey) and the inferred (white) word vector
    representations.} 
  \label{fig:word-graph}
\end{figure}

Let $V = \{w_1, \ldots ,w_n\}$ be a \textbf{vocabulary}, i.e, the set of word types, and $\Omega$ be an \textbf{ontology} that encodes semantic relations between words in $V$.
We represent $\Omega$ as an undirected graph $(V,E)$
with one vertex for each word type and
edges $(w_i, w_j) \in E \subseteq V \times V$ indicating a semantic relationship
of interest. These relations differ for different semantic lexicons and
are described later (\S\ref{sec:lexicons}).

The matrix $\hat{Q}$ will be the collection of vector representations
$\hat{q}_i \in \mathbb{R}^d$, 
for each $w_i \in V$, learned using a standard data-driven technique, where
$d$ is the length of the word vectors.
Our objective is to learn the matrix $Q = ( q_1, \ldots, q_n )$ such that the columns are both close (under a distance metric) to their counterparts in $\hat{Q}$ \emph{and} to adjacent vertices in
 $\Omega$.
Figure~\ref{fig:word-graph} shows a small word graph with such edge
connections; white nodes are labeled with the $Q$ vectors to be
retrofitted (and correspond to $V_\Omega$); shaded nodes are labeled with
the corresponding vectors in $\hat{Q}$, which are observed.  The graph
can be interpreted as a Markov random field
\cite{kindermann80mrf}.

The distance between a pair of vectors is defined to be the Euclidean distance. Since we want the inferred word vector to be
close to the observed value $\hat{q}_i$ and close to its neighbors
$q_j, \forall j$ such that $(i, j) \in E$, the objective to
be minimized becomes:
\begin{equation*}
  \displaystyle \Psi(Q) = \sum_{i=1}^n \left[ \alpha_i \lVert q_i - \hat{q_i} \rVert^2 + \sum_{(i,j) \in E} \beta_{ij} \lVert q_i - q_j \rVert^2 \right]
\end{equation*}
where $\alpha$ and $\beta$ values control the relative strengths of
associations (more details in \S\ref{sec:exp-post-proc}).

In this case, we first train the word vectors independent of the information in the semantic lexicons
and then retrofit them. $\Psi$ is convex in $Q$ and its  solution can be found by solving a
 system of linear equations. To do so, we use an efficient iterative
updating method
\cite{Bengio+al-ssl-2006,Subramanya:2010:EGS:1870658.1870675,das-petrov:2011:ACL-HLT2011,das-smith:2011:ACL-HLT2011}.
The vectors in $Q$  are initialized to be equal to the
vectors in $\hat{Q}$.
We take the first derivative of $\Psi$ with respect to one $q_i$
vector, and by equating it to zero arrive at the following online update:
\begin{equation}
  \label{equ:update-vector}
q_i = \frac{\sum_{j : (i,j) \in E} \beta_{ij} q_j + \alpha_i
  \hat{q_i}}{\sum_{j : (i,j) \in E} \beta_{ij} + \alpha_i}
\end{equation}
In practice, running this procedure for $10$ iterations converges to
changes in Euclidean distance of adjacent vertices of less than
$10^{-2}$. 
The retrofitting approach described above is modular; it can be applied to word vector
representations obtained from any model as the updates in
Eq.~\ref{equ:update-vector} are agnostic to the original vector training
model objective.

\paragraph{Semantic Lexicons during Learning.}
Our proposed approach is reminiscent of recent work on improving word vectors using lexical
resources \cite{Yu:2014,yan-ecml14,xu2014rc} which alters the learning
objective of the original vector training model with a prior
(or a regularizer) that encourages semantically related vectors
(in $\Omega$) to be close together, except that our technique is
applied as a second stage of learning. We describe the prior approach here since it will serve as a baseline.
Here semantic lexicons play the role of a prior on $Q$ which we define as follows:
\begin{equation}
  \label{equ:prior}
  p(Q) \propto \text{exp}\left(-\gamma\sum_{i=1}^{n}\sum_{j : (i,j) \in E}\beta_{ij} \lVert q_{i}-q_{j} \rVert^{2}\right)
\end{equation}
Here, $\gamma$ is a hyperparameter that controls the strength of the prior.
As in the retrofitting objective, this prior on the word
vector parameters forces words connected in the lexicon to have close
vector representations as did $\Psi(Q)$ (with the role of $\hat{Q}$ being played
by cross entropy of the empirical distribution).

This prior can be incorporated during learning through maximum a posteriori (MAP)
estimation.
Since there is no closed form solution
of the estimate, we consider two iterative procedures.
In the first, we use the sum of gradients of the
log-likelihood (given by the extant vector learning model) and the
log-prior (from Eq.~\ref{equ:prior}), with respect to $Q$ for learning.
Since computing the gradient of Eq.~\ref{equ:prior}
has linear runtime in the vocabulary size $n$, we use lazy updates~\cite{Carpenter08lazysparse} for every
$k$ words during training.
We call this the \textbf{lazy} method of MAP.
The second technique applies stochastic gradient ascent to the
log-likelihood, and after every $k$ words applies the update in
Eq.~\ref{equ:update-vector}.
We call this the \textbf{periodic} method.
We later experimentally compare these methods against retrofitting (\S\ref{sec:improve-tr}).

\section{Word Vector Representations}
\label{sec:vectors}

We now describe the various publicly available pre-trained English
word vectors
on which we will test the applicability of the retrofitting model.
These vectors have been chosen to have a balanced mix between
large and small amounts of unlabeled text as well as
between neural and spectral methods of training word vectors.


\paragraph{Glove Vectors.} Global vectors for word representations~\cite{glove:2014} are trained on aggregated global word-word co-occurrence statistics from a corpus, and the resulting representations show interesting linear substructures of the word vector space.
These vectors were trained on 6 billion words from Wikipedia and English Gigaword and
are of length $300$.\footnote{\url{http://www-nlp.stanford.edu/projects/glove/}}

\paragraph{Skip-Gram Vectors (SG).}
The \texttt{word2vec} tool~\cite{mikolov2013efficient} is  fast and
currently in wide use.
In this model, each word's Huffman code is used as an input to a log-linear classifier
with a continuous projection layer and words within a given context window are predicted.
The available vectors are trained on 100 billion words of Google news dataset and are of length $300$.\footnote{\url{https://code.google.com/p/word2vec}}

\paragraph{Global Context Vectors (GC).}
These vectors are learned using a recursive neural network that
incorporates both local and global (document-level) context features \cite{huang2012improving}.
These vectors were trained on the first 1 billion words of English
Wikipedia and are of length
$50$.\footnote{\url{http://nlp.stanford.edu/~socherr/ACL2012_wordVectorsTextFile.zip}} 

\paragraph{Multilingual Vectors (Multi).}
\newcite{faruqui-dyer:2014:EACL2014} learned vectors by first
performing SVD on text in different languages, then applying canonical
correlation analysis (CCA) on pairs of vectors for words that align in
parallel corpora. The monolingual vectors were trained on WMT-2011 news corpus
for English, French, German and Spanish. We use the Enligsh word vectors
projected in the common English--German space.
The monolingual English WMT corpus had 360 million words and the trained
vectors are of length
$512$.\footnote{\url{http://cs.cmu.edu/~mfaruqui/soft.html}}

\section{Semantic Lexicons}
\label{sec:lexicons}

\begin{table}[!t]
  \centering
  \small
  \begin{tabular}{|l|r|r|}
  \hline
Lexicon & Words & Edges \\
\hline
PPDB & 102,902 & 374,555 \\
WordNet$_{syn}$ & 148,730 & 304,856\\
WordNet$_{all}$ & 148,730 &  934,705 \\
FrameNet & 10,822 & 417,456\\
\hline
\end{tabular}
   \caption{Approximate size of the graphs obtained from different
     lexicons.} 
  \label{tab:lexicons}
\end{table}

We use three different semantic lexicons to evaluate their utility in improving
the word vectors. We include both manually and
automatically created lexicons. Table~\ref{tab:lexicons} shows the size of the
graphs obtained from these lexicons.

\paragraph{PPDB.}
The paraphrase database
\cite{ganitkevitch2013ppdb} is a semantic lexicon containing more
than 220 million paraphrase pairs of English.\footnote{\url{http://www.cis.upenn.edu/~ccb/ppdb}}  Of these, 8 million are
lexical (single word to single word) paraphrases. The key intuition
behind the acquisition of its lexical paraphrases is that two words in one
language that align, in parallel text, to the same word in a different
language, should be synonymous. For example, if the words
\textit{jailed} and \textit{imprisoned} are translated as the same word in
another language, it may be reasonable to assume they have the same
meaning.
In our experiments, we instantiate an edge in $E$ for each
lexical paraphrase in PPDB. The lexical paraphrase dataset comes in
different sizes ranging from S to XXXL, in decreasing order of
paraphrasing confidence and increasing order of size.  We chose XL
for our experiments. 
We want to give higher edge weights ($\alpha_i$) connecting the retrofitted word vectors ($q$) to the purely distributional word vectors ($\hat{q}$) than to edges connecting the retrofitted vectors to each other ($\beta_{ij}$), so all $\alpha_i$ are set to 1 and $\beta_{ij}$ to be $\mathrm{degree}(i)^{-1}$ (with $i$ being the node the update is being applied to).\footnote{In principle, these hyperparameters can be tuned to optimize performance on a particular task, which we leave for future work.}

\paragraph{WordNet.}
WordNet~\cite{miller:1995} is a large human-constructed semantic lexicon of English words.
It groups English words into sets of synonyms called synsets, provides short, general definitions, and records the various semantic relations between synsets.
This database is structured in a graph particularly suitable for our
task because it explicitly relates concepts with semantically aligned
relations such as hypernyms and hyponyms. For example, the word
\textit{dog} is a synonym of \textit{canine}, 
a hypernym of \textit{puppy}
and a hyponym of \textit{animal}. We perform two different experiments
with WordNet: (1) connecting a word only to synonyms, and (2) connecting a word to synonyms, hypernyms and hyponyms.
We refer to these two graphs as WN$_{\mathit{syn}}$ and
WN$_{\mathit{all}}$, respectively. In both settings, all $\alpha_i$ are set to 1 and
$\beta_{ij}$ to be $\mathrm{degree}(i)^{-1}$.

\paragraph{FrameNet.}
FrameNet~\cite{Baker:1998:BFP:980845.980860,fillmore-ua-2003} is a rich linguistic resource containing  information about
lexical and predicate-argument semantics in English. 
Frames can be realized on the surface by many different word types,
which suggests that the word types evoking the same frame should be semantically
related.
 For example, the frame Cause\_change\_of\_position\_on\_a\_scale
is associated with \emph{push}, \emph{raise}, and \emph{growth} (among
many others).
In our use of FrameNet, two words that group
together with any frame are given an edge in $E$.
We refer to this graph as FN.
All $\alpha_i$ are set to 1 and  $\beta_{ij}$ to be $\mathrm{degree}(i)^{-1}$.

\section{Evaluation Benchmarks}
\label{sec:eval}

We evaluate the quality of our word vector representations on tasks that
test how well they capture both semantic and syntactic aspects of the representations
along with an extrinsic sentiment analysis task.

\paragraph{Word Similarity.}
We evaluate our word representations on a variety of different benchmarks that
have been widely used to measure word similarity. The first one is the \textbf{WS-353}
dataset~\cite{citeulike:379845} containing 353 pairs of English words that have been
assigned similarity ratings by humans. The second benchmark is the \textbf{RG-65}
\cite{Rubenstein:1965:CCS:365628.365657} dataset that contain 65 pairs of nouns.
Since the commonly used word similarity datasets contain a small number of word pairs we
also use the \textbf{MEN} dataset~\cite{bruni:2012} of 3,000 word
pairs sampled from words
that occur at least 700 times in a large web corpus.
We calculate cosine similarity between the vectors of two words
forming a test item, and report Spearman's rank correlation coefficient
\cite{citeulike:8703893} between the rankings produced by our model against the
human rankings.

\paragraph{Syntactic Relations (SYN-REL).}
\newcite{mikolov-yih-zweig:2013:NAACL} present a syntactic relation
dataset composed of analogous word pairs.
It contains pairs of tuples of word relations that follow a common
syntactic relation. For example, given \textit{walking} and
\textit{walked}, the words are differently inflected forms of the same
verb.
There are nine  different kinds of relations and
overall there are 10,675 syntactic pairs of word tuples.
The task is to find a word \emph{d} that best fits the following relationship:
``\emph{a} is to \emph{b} as \emph{c}  is to \emph{d},'' given
\emph{a}, \emph{b}, and \emph{c}.
We use the vector offset method \cite{mikolov2013efficient,levy-goldberg:2014:W14-16},
computing  $q = q_a - q_b + q_c$ and returning the
vector from $Q$ which has the highest cosine similarity to $q$.

\paragraph{Synonym Selection (TOEFL).}
The TOEFL synonym selection task is to select the semantically closest word
to a target from a list of four candidates~\cite{landauer:1997}.
The dataset contains 80 such questions. An example is
``\emph{rug} $\rightarrow$ $\{$\emph{sofa}, \emph{ottoman},
\emph{carpet}, \emph{hallway}$\}$'', with \emph{carpet} being the most synonym-like candidate to the target.

\paragraph{Sentiment Analysis (SA).}
\newcite{Socher-etal:2013} created a treebank containing sentences
annotated with fine-grained sentiment labels on phrases and sentences
from movie review excerpts.
The coarse-grained treebank of positive and negative
classes has been split into training, development, and test datasets
containing 6,920, 872, and
1,821 sentences, respectively. We train an $\ell_2$-regularized  logistic
regression classifier on the average of the word
vectors of a given sentence to predict the coarse-grained sentiment
tag at the sentence level, and report the test-set accuracy of the classifier.

\section{Experiments}
\label{sec:expts}

We first show experiments measuring improvements from the retrofitting method
(\S\ref{sec:exp-post-proc}), followed by  comparisons to using
lexicons during MAP learning (\S\ref{sec:improve-tr}) and other published
methods (\S\ref{sec:compare}). We then test how well
retrofitting generalizes to other languages (\S\ref{sec:multilingual}).

\begin{table*}[!th]
  \centering
  \small
  \begin{tabular}{|l||r|r|r||r|r||r|}
  \hline
Lexicon & MEN-3k & RG-65 & WS-353 & TOEFL & SYN-REL & SA \\
\hline\hline
Glove & 73.7 & 76.7 & 60.5 & 89.7 & 67.0 & 79.6 \\
\hline
~+PPDB & 1.4 & 2.9 & --1.2 & \textbf{5.1} & --0.4 & \textbf{1.6} \\
~+WN$_{\mathit{syn}}$ & 0.0 & 2.7 & 0.5 & \textbf{5.1} & --12.4 & 0.7 \\
~+WN$_{\mathit{all}}$ & \textbf{2.2} & \textbf{7.5} & \textbf{0.7} & 2.6 & --8.4  & 0.5 \\
~+FN & --3.6 & --1.0 & --5.3 & 2.6 & --7.0 & 0.0 \\
\hline\hline
SG & 67.8 & 72.8 & 65.6 & 85.3 & 73.9 & 81.2\\
\hline
~+PPDB & \textbf{5.4} & 3.5 & \textbf{4.4} & \textbf{10.7} & --2.3 & \textbf{0.9} \\
~+WN$_{\mathit{syn}}$ & 0.7 & 3.9 & 0.0 & 9.3 & --13.6 & 0.7 \\
~+WN$_{\mathit{all}}$ & 2.5 & \textbf{5.0} & 1.9 & 9.3 & --10.7 & --0.3 \\
~+FN & --3.2 & 2.6 & --4.9 & 1.3 & --7.3 & 0.5 \\
\hline\hline
GC & 31.3	& 62.8	& 62.3 & 60.8 & 10.9 & 67.8 \\
\hline
~+PPDB & \textbf{7.0} & 6.1 & 2.0 & \textbf{13.1} & \textbf{5.3} & \textbf{1.1} \\
~+WN$_{\mathit{syn}}$ & 3.6 & 6.4 & 0.6 & 7.3 & --1.7 & 0.0 \\
~+WN$_{\mathit{all}}$ & 6.7 & \textbf{10.2} & \textbf{2.3} & 4.4 & --0.6 & 0.2 \\
~+FN & 1.8 & 4.0 & 0.0 & 4.4 & --0.6 & 0.2 \\
\hline\hline
Multi&  75.8 & 75.5 & 68.1 & 84.0 & 45.5 & 81.0 \\
\hline
~+PPDB & \textbf{3.8} & 4.0 & \textbf{6.0} & \textbf{12.0} & \textbf{4.3} & 0.6 \\
~+WN$_{\mathit{syn}}$ & 1.2 & 0.2 & 2.2 & 6.6 & --12.3 & \textbf{1.4} \\
~+WN$_{\mathit{all}}$ & 2.9 & \textbf{8.5} & 4.3 & 6.6 & --10.6 & \textbf{1.4} \\
~+FN & 1.8 & 4.0 & 0.0 & 4.4 & --0.6 & 0.2 \\
\hline
\end{tabular}
\caption{Absolute performance changes with retrofitting. Spearman's correlation
(3 left columns) and accuracy (3 right columns) on different tasks.
Higher scores are always better.  Bold indicates greatest
improvement for a vector type.}
  \label{tab:retrofit}
\end{table*}

\subsection{Retrofitting}
\label{sec:exp-post-proc}

We use Eq.~\ref{equ:update-vector} to retrofit word vectors
(\S\ref{sec:vectors}) using graphs derived from
semantic lexicons (\S\ref{sec:lexicons}).

\paragraph{Results.} Table~\ref{tab:retrofit} shows the absolute
changes in performance on different tasks (as columns) with different semantic lexicons (as rows).
All of the lexicons offer high improvements on the word similarity tasks
(the first three columns). On the TOEFL task,
we observe large improvements of the order of $10$ absolute points
in accuracy for all lexicons except for FrameNet. FrameNet's performance is
weaker, in some cases leading to worse performance (e.g., with Glove and
SG vectors).
For the extrinsic sentiment analysis task, we observe
improvements using all the lexicons and gain $1.4\%$ (absolute)
in accuracy for the Multi
vectors over the baseline. This increase is statistically significant
($p<0.01$, McNemar).

We observe improvements over Glove and SG vectors, which were trained on billions of tokens on all tasks except for SYN-REL.
For stronger baselines (Glove and Multi) we observe
smaller improvements as compared to lower baseline scores (SG and GC).
We believe that FrameNet does not perform as well as the other
lexicons because its frames group words based on very abstract concepts;
often words with seemingly distantly related meanings (e.g., \emph{push} and \emph{growth}) can evoke the same frame.
Interestingly, we almost never improve on the SYN-REL task, especially with
higher baselines, this can be attributed to the fact that SYN-REL is
inherently a syntactic task and during retrofitting we are
incorporating additional semantic
information in the vectors.
In summary, we find that PPDB gives the best improvement maximum number
of times aggreagted over different vetor types,
closely followed by WN$_{\mathit{all}}$,
and retrofitting gives gains across tasks and vectors.
An ensemble lexicon, in which the graph is the union of the WN$_{\mathit{all}}$ and PPDB lexicons, on average performed slightly worse than PPDB; we omit those results here for brevity.

\subsection{Semantic Lexicons during Learning}
\label{sec:improve-tr}

\begin{table*}[!th]
  \centering
  \small
  \begin{tabular}{|l|r||r|r|r||r|r||r|}
  \hline
Method & $k$, $\gamma$ & MEN-3k & RG-65 & WS-353 & TOEFL & SYN-REL & SA \\
\hline
 LBL (Baseline) & $k=\infty$, $\gamma=0$& 58.0 & 42.7  & 53.6 & 66.7 & 31.5 & 72.5 \\
\hline
\multirow{3}{*}{\textbf{LBL + Lazy}} & $\gamma=1$& --0.4 & 4.2 & 0.6 & --0.1 & 0.6 & 1.2\\
 & $\gamma=0.1$& 0.7 & 8.1 & 0.4  & --1.4 & 0.7 & 0.8\\
 & $\gamma=0.01$& 0.7 & 9.5 & 1.7  & 2.6 & 1.9 &  0.4\\
\hline
\multirow{3}{*}{\textbf{LBL + Periodic}} & $k=100$M& 3.8 & 18.4 & 3.6  & 12.0 &
4.8 & 1.3\\
 & $k=50$M & 3.4 & \textbf{19.5} & 4.4  & 18.6 & 0.6 &  \textbf{1.9} \\
 & $k=25$M & 0.5 & 18.1 & 2.7  & \textbf{21.3} & --3.7 & 0.8\\
 \hline
 \textbf{LBL + Retrofitting} & -- & \textbf{5.7} & 15.6 & \textbf{5.5} & 18.6 & \textbf{14.7} & 0.9\\
 \hline
\end{tabular}
   \caption{Absolute performance changes for including PPDB information while training LBL vectors. Spearman's correlation (3 left columns) and accuracy (3 right columns) on different tasks. Bold
   indicates greatest improvement. 
     }
  \label{tab:lbl-tr}
\end{table*}


To incorporate lexicon information during training, and compare its
performance against retrofitting,
we train log-bilinear (LBL) vectors~\cite{MnihTeh2012}.
These vectors are trained to optimize the log-likelihood of a language model
which predicts a word token $w$'s
vector given the set of words in its context ($h$), also represented as vectors:
\begin{equation}
  \label{equ:prob-lbl}
  p(w \mid h;Q) \propto \exp \left(  \sum_{i \in h} q_i^\top q_j + b_j\right)
\end{equation}
We optimize the above likelihood combined with the prior defined in Eq.~\ref{equ:prior}
using the {lazy} and {periodic} techniques described in \S\ref{sec:framework}.
Since it is costly to compute the partition function over the whole vocabulary,
we use \textit{noise constrastive estimation} (NCE) to estimate the parameters
of the model~\cite{MnihTeh2012} using AdaGrad~\cite{Duchi:EECS-2010-24} with a
learning rate of $0.05$.

We train vectors of length $100$ on the WMT-2011 news corpus, which contains $360$ million words, and use PPDB as the semantic lexicon as it performed reasonably well in the retrofitting experiments (\S\ref{sec:exp-post-proc}).
 For the {lazy} method
we update with respect to the prior every $k=$ 100,000
words\footnote{$k =$ 10,000 or 50,000 yielded similar results.}
and test for different values of
prior strength $\gamma \in \{1, 0.1, 0.01\}$.
For the {periodic} method, we update the word vectors using Eq.~\ref{equ:update-vector}
every $k \in \{25, 50, 100\}$ million words.

\paragraph{Results.} See Table~\ref{tab:lbl-tr}. For {lazy},
$\gamma = 0.01$ performs best, but the method is in most cases not
highly sensitive to $\gamma$'s value.
For \textbf{periodic}, which overall leads to greater improvements
over the baseline than \textbf{lazy}, $k=50$M performs best,
although all other values of $k$
also outperform the the baseline.
Retrofitting, which can be applied to any word vectors, regardless of
how they are trained, is competitive and sometimes better.

\subsection{Comparisons to Prior Work}
\label{sec:compare}

\begin{table*}[!th]
  \centering
  \small
  \begin{tabular}{|l|l||r|r|r||r|r||r|}
  \hline
Corpus & Vector Training & MEN-3k & RG-65 & WS-353 & TOEFL & SYN-REL & SA \\
\hline
\multirow{3}{*}{WMT-11} & CBOW & 55.2 & 44.8 & 54.7 & 73.3 & 40.8 & 74.1\\
& \newcite{Yu:2014} & 50.1 & 47.1  & 53.7 & 61.3 & 29.9 & 71.5\\
& CBOW + Retrofitting &\textbf{60.5} & \textbf{57.7} & \textbf{58.4} & \textbf{81.3}  & \textbf{52.5} & \textbf{75.7}\\
\hline \hline
\multirow{3}{*}{Wikipedia} & SG & \textbf{76.1} & 66.7 & \textbf{68.6} & 72.0 & 40.3 & 73.1 \\
& \newcite{xu2014rc} & -- & -- & 68.3 & -- & 44.4 & -- \\
& SG + Retrofitting & 65.7 & \textbf{73.9} & 67.5 & \textbf{86.0} & \textbf{49.9} & \textbf{74.6}\\
\hline
\end{tabular}
\caption{Comparison of retrofitting for semantic enrichment against
 \newcite{Yu:2014}, \newcite{xu2014rc}. Spearman's correlation (3 left columns) and accuracy (3 right columns) on different tasks.}
  \label{tab:compare}
\end{table*}

Two previous models \cite{Yu:2014,xu2014rc} have shown that the quality of word
vectors obtained using \texttt{word2vec} tool can be improved by using semantic
knowledge from lexicons. Both these models use constraints among words as a
regularization term on the training objective during training, and their methods
can only be applied for improving the quality of SG and CBOW vectors produced
by the \texttt{word2vec} tool. We compared the quality of our vectors against
each of these.

\paragraph{\newcite{Yu:2014}.} We train word vectors using their joint model training code\footnote{\url{https://github.com/Gorov/JointRCM}} while using exactly the same training settings as specified in their best model: CBOW, vector length 100 and PPDB for enrichment. The results are shown in the top half of Table~\ref{tab:compare} where our model consistently outperforms the baseline and their model.

\paragraph{\newcite{xu2014rc}.} This model extracts categorical and relational knowledge among words from Freebase\footnote{\url{https://www.freebase.com}} and uses it as a constraint while training. Unfortunately, neither their word embeddings nor model training code is publicly available, so we train the SG model by using exactly the same settings as described in their system (vector length 300) and on the same corpus: monolingual English Wikipedia text.\footnote{\url{http://mattmahoney.net/dc/enwik9.zip}} We compare the performance of our retrofitting vectors on the {SYN-REL} and {WS-353} task against the best model\footnote{Their best model is named ``RC-NET'' in their paper.} reported in their paper. As shown in the lower half of Table~\ref{tab:compare}, our model outperforms their model by an absolute $5.5$ points absolute on the {SYN-REL} task, but a slightly inferior score on the
{WS-353} task.

\subsection{Multilingual Evaluation}
\label{sec:multilingual}

We tested our method on three additional languages: German,
French, and Spanish. We used the Universal WordNet
\cite{deMeloWeikum2009}, an automatically constructed multilingual lexical knowledge base based on
WordNet.\footnote{\url{http://www.mpi-inf.mpg.de/yago-naga/uwn}}
It contains words connected via different lexical relations to other words both
within and across languages. We construct separate graphs for different
languages (i.e., only linking words to other words in the same
language) and apply retrofitting to each.
Since not many word similarity evaluation benchmarks are available for
 languages other than English, we tested our baseline and improved vectors on one benchmark per language.

We used RG-65 \cite{Gurevych:2005:USC:2145899.2145986}, RG-65 \cite{Joubarne:2011:CSS:2018192.2018218}
and MC-30 \cite{Hassan:2009:CSR:1699648.1699665} for German, French and Spanish, respectively.\footnote{These benchmarks were created by translating the corresponding English benchmarks word by word manually.}
We trained SG vectors for each language of length 300 on a corpus of 1 billion tokens, each extracted from Wikipedia,
and evaluate them on word similarity on the benchmarks before and after retrofitting.  Table~\ref{tab:multi-lang} shows that we obtain
high improvements which strongly indicates that our method generalizes
across these languages.

\begin{table}[!t]
  \centering
  \small
  \begin{tabular}{|l|l|r|r|}
  \hline
 Language & Task & SG & Retrofitted SG \\
\hline
German & RG-65 &  53.4 & \textbf{60.3} \\
French & RG-65 &  46.7 & \textbf{60.6} \\
Spanish & MC-30 & 54.0 & \textbf{59.1}\\
\hline
\end{tabular}
   \caption{Spearman's correlation for word similarity evaluation using the
  using original and retrofitted SG vectors.}
  \label{tab:multi-lang}
\end{table}

\section{Further Analysis}
\label{sec:analysis}

\begin{figure}[!tb]
  \centering
  \includegraphics[width=0.9\columnwidth]{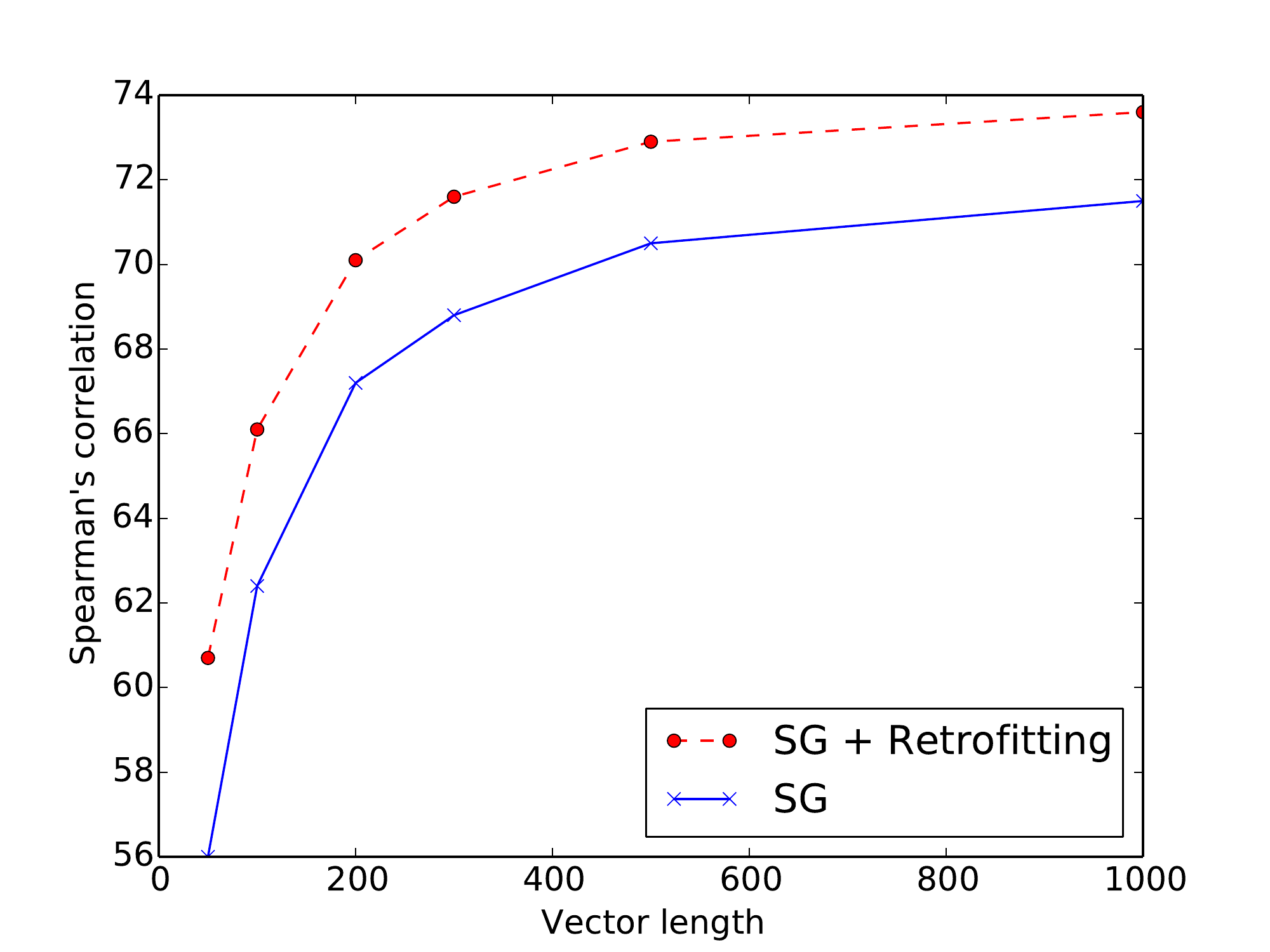}
  \caption{Spearman's correlation
    on the {MEN} word similarity task, before and after retrofitting.}
  \label{fig:len}
\end{figure}

\begin{figure*}[!tbh]
  \centering
  \includegraphics[width=1.8\columnwidth]{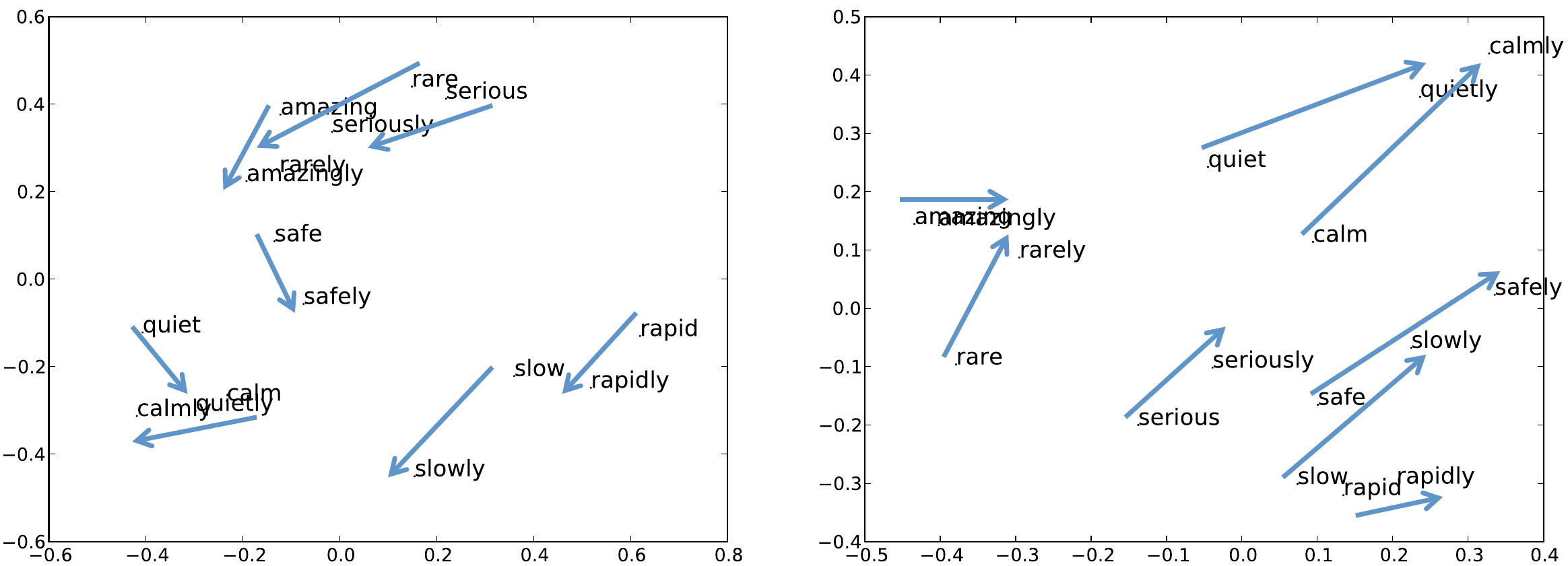}
  \caption{Two-dimensional PCA projections of 100-dimensional {SG}
    vector pairs holding the ``adjective to adverb'' relation, before (left) and after (right) retrofitting.}
  \label{fig:analogy}
\end{figure*}

\paragraph{Retrofitting vs.~vector length.} With more dimensions, word
vectors might be able to capture higher orders of semantic information
and retrofitting might be less helpful.
We train SG vectors on 1 billion English tokens for vector lengths ranging from 50 to 1,000 and evaluate on the
{MEN} word similarity task. We retrofit these vectors to PPDB (\S\ref{sec:lexicons}) and evaluate those on the same task. Figure~\ref{fig:len} shows consistent improvement in vector quality across different vector lengths.

\paragraph{Visualization.} We randomly select eight word pairs that
have the ``adjective to adverb'' relation from the SYN-REL task
(\S\ref{sec:eval}). We then take a two-dimensional PCA projection of
the 100-dimensional SG word vectors and plot them in
$\mathbb{R}^2$. In Figure~\ref{fig:analogy} we plot these projections before (left)
and after (right) retrofitting.
  It can be seen that in the first case the direction of the analogy vectors is
  not consistent, but after retrofitting all the analogy vectors are aligned in
  the same direction.

\section{Related Work}

The use of lexical semantic information in training word vectors has been limited.
Recently, word similarity knowledge \cite{Yu:2014,fried2014incorporating} and word relational knowledge
\cite{xu2014rc,yan-ecml14} have been used to improve the
\texttt{word2vec} embeddings in a joint training model similar to our regularization
approach.
In latent semantic analysis, the word cooccurrence matrix can be
constructed to incorporate relational
information like antonym specific polarity induction~\cite{Yih:2012} and multi-relational latent semantic analysis~\cite{chang-yih-meek:2013:EMNLP}.


The approach we propose is conceptually similar to previous work that uses graph
structures to propagate information among semantic concepts~\cite{Zhu:2005:SLG:1104523,10.1109/TPAMI.2007.70765}.
Graph-based belief propagation has also been used to induce POS tags
\cite{Subramanya:2010:EGS:1870658.1870675,das-petrov:2011:ACL-HLT2011}
and semantic frame associations \cite{das-smith:2011:ACL-HLT2011}.
In those efforts, labels for unknown words were inferred using a method
similar to ours.
Broadly, graph-based
semi-supervised
learning~\cite{Zhu:2005:SLG:1104523,Talukdar:2010:EGS:1858681.1858830}
has been applied to machine translation \cite{Alexandrescu:2009:GLS:1620754.1620772},
unsupervised semantic role induction \cite{Lang:2011:USR:2145432.2145571},
semantic document modeling~\cite{schuhmacher:2014},
language generation~\cite{Krahmer:2003:GGR:778822.778825}
and sentiment analysis~\cite{Goldberg:2006:SSA:1654758.1654769}.

\section{Conclusion}

We have proposed a simple and effective
method named \textbf{retrofitting} to improve
word vectors using word relation knowledge found in semantic lexicons. 
Retrofitting is used as
a post-processing step to improve vector quality and is more modular
than other approaches that use semantic information while training.
It can be applied to vectors obtained from any word vector training
method.  Our experiments explored the method's performance across
tasks, semantic lexicons, and languages and showed that it outperforms
existing alternatives.
The retrofitting tool is available at:
\url{https://github.com/mfaruqui/retrofitting}.

\section*{Acknowledgements}

This research was supported in part by the National Science Foundation under grants IIS-1143703,
IIS-1147810, and IIS-1251131; by IARPA via Department of Interior National Business Center
(DoI/NBC) contract number D12PC00337; and by DARPA under grant FA87501220342. Part of
the computational work was carried out on resources provided by the Pittsburgh Supercomputing
Center. The U.S. Government is authorized to reproduce and distribute reprints for Governmental
purposes notwithstanding any copyright annotation thereon. Disclaimer: the views and conclusions
contained herein are those of the authors and should not be interpreted as necessarily representing
the official policies or endorsements, either expressed or implied, of IARPA, DoI/NBC, DARPA, or
the U.S. Government.

\bibliography{references}

\begin{thebibliography}{}

\bibitem[\protect\citename{Agirre \bgroup et al.\egroup
  }2009]{Agirre:2009:SSR:1620754.1620758}
Eneko Agirre, Enrique Alfonseca, Keith Hall, Jana Kravalova, Marius Pa\c{s}ca,
  and Aitor Soroa.
\newblock 2009.
\newblock A study on similarity and relatedness using distributional and
  wordnet-based approaches.
\newblock In {\em Proceedings of NAACL}.

\bibitem[\protect\citename{Alexandrescu and
  Kirchhoff}2009]{Alexandrescu:2009:GLS:1620754.1620772}
Andrei Alexandrescu and Katrin Kirchhoff.
\newblock 2009.
\newblock Graph-based learning for statistical machine translation.
\newblock NAACL '09.

\bibitem[\protect\citename{Baker \bgroup et al.\egroup
  }1998]{Baker:1998:BFP:980845.980860}
Collin~F. Baker, Charles~J. Fillmore, and John~B. Lowe.
\newblock 1998.
\newblock The berkeley framenet project.
\newblock ACL '98.

\bibitem[\protect\citename{Bengio \bgroup et al.\egroup
  }2006]{Bengio+al-ssl-2006}
Yoshua Bengio, Olivier Delalleau, and Nicolas {Le Roux}.
\newblock 2006.
\newblock Label propagation and quadratic criterion.
\newblock In {\em Semi-Supervised Learning}.

\bibitem[\protect\citename{Bian \bgroup et al.\egroup }2014]{yan-ecml14}
Jiang Bian, Bin Gao, and Tie-Yan Liu.
\newblock 2014.
\newblock Knowledge-powered deep learning for word embedding.
\newblock In {\em Machine Learning and Knowledge Discovery in Databases}.

\bibitem[\protect\citename{Bruni \bgroup et al.\egroup }2012]{bruni:2012}
Elia Bruni, Gemma Boleda, Marco Baroni, and Nam-Khanh Tran.
\newblock 2012.
\newblock Distributional semantics in technicolor.
\newblock In {\em Proceedings of ACL}.

\bibitem[\protect\citename{Carpenter}2008]{Carpenter08lazysparse}
Bob Carpenter.
\newblock 2008.
\newblock Lazy sparse stochastic gradient descent for regularized multinomial
  logistic regression.

\bibitem[\protect\citename{Chang \bgroup et al.\egroup
  }2013]{chang-yih-meek:2013:EMNLP}
Kai-Wei Chang, Wen-tau Yih, and Christopher Meek.
\newblock 2013.
\newblock Multi-relational latent semantic analysis.
\newblock In {\em Proceedings of EMNLP}.

\bibitem[\protect\citename{Collobert and
  Weston}2008]{Collobert:2008:UAN:1390156.1390177}
Ronan Collobert and Jason Weston.
\newblock 2008.
\newblock A unified architecture for natural language processing: deep neural
  networks with multitask learning.
\newblock In {\em Proceedings of ICML}.

\bibitem[\protect\citename{Culp and Michailidis}2008]{10.1109/TPAMI.2007.70765}
Mark Culp and George Michailidis.
\newblock 2008.
\newblock Graph-based semisupervised learning.
\newblock {\em IEEE Transactions on Pattern Analysis and Machine Intelligence}.

\bibitem[\protect\citename{Das and Petrov}2011]{das-petrov:2011:ACL-HLT2011}
Dipanjan Das and Slav Petrov.
\newblock 2011.
\newblock Unsupervised part-of-speech tagging with bilingual graph-based
  projections.
\newblock In {\em Proc. of ACL}.

\bibitem[\protect\citename{Das and Smith}2011]{das-smith:2011:ACL-HLT2011}
Dipanjan Das and Noah~A. Smith.
\newblock 2011.
\newblock Semi-supervised frame-semantic parsing for unknown predicates.
\newblock In {\em Proc. of ACL}.

\bibitem[\protect\citename{de Melo and Weikum}2009]{deMeloWeikum2009}
Gerard de~Melo and Gerhard Weikum.
\newblock 2009.
\newblock Towards a universal wordnet by learning from combined evidence.
\newblock In {\em Proceedings of CIKM}.

\bibitem[\protect\citename{Deerwester \bgroup et al.\egroup
  }1990]{deerwester-90}
S.~C. Deerwester, S.~T. Dumais, T.~K. Landauer, G.~W. Furnas, and R.~A.
  Harshman.
\newblock 1990.
\newblock Indexing by latent semantic analysis.
\newblock {\em Journal of the American Society for Information Science}.

\bibitem[\protect\citename{Duchi \bgroup et al.\egroup
  }2010]{Duchi:EECS-2010-24}
John Duchi, Elad Hazan, and Yoram Singer.
\newblock 2010.
\newblock Adaptive subgradient methods for online learning and stochastic
  optimization.
\newblock Technical Report UCB/EECS-2010-24, Mar.

\bibitem[\protect\citename{Faruqui and Dyer}2014]{faruqui-dyer:2014:EACL2014}
Manaal Faruqui and Chris Dyer.
\newblock 2014.
\newblock Improving vector space word representations using multilingual
  correlation.
\newblock In {\em Proceedings of EACL}.

\bibitem[\protect\citename{Fillmore \bgroup et al.\egroup
  }2003]{fillmore-ua-2003}
Charles Fillmore, Christopher Johnson, and Miriam Petruck.
\newblock 2003.
\newblock {\em International Journal of Lexicography}.

\bibitem[\protect\citename{Finkelstein \bgroup et al.\egroup
  }2001]{citeulike:379845}
Lev Finkelstein, Evgeniy Gabrilovich, Yossi Matias, Ehud Rivlin, Zach Solan,
  Gadi Wolfman, and Eytan Ruppin.
\newblock 2001.
\newblock {Placing search in context: the concept revisited}.
\newblock In {\em WWW}, New York, NY, USA.

\bibitem[\protect\citename{Fried and Duh}2014]{fried2014incorporating}
Daniel Fried and Kevin Duh.
\newblock 2014.
\newblock Incorporating both distributional and relational semantics in word
  representations.
\newblock {\em arXiv preprint arXiv:1412.4369}.

\bibitem[\protect\citename{Ganitkevitch \bgroup et al.\egroup
  }2013]{ganitkevitch2013ppdb}
Juri Ganitkevitch, Benjamin {Van Durme}, and Chris Callison-Burch.
\newblock 2013.
\newblock {PPDB}: The paraphrase database.
\newblock In {\em Proceedings of NAACL}.

\bibitem[\protect\citename{Goldberg and
  Zhu}2006]{Goldberg:2006:SSA:1654758.1654769}
Andrew~B. Goldberg and Xiaojin Zhu.
\newblock 2006.
\newblock Seeing stars when there aren't many stars: Graph-based
  semi-supervised learning for sentiment categorization.
\newblock TextGraphs-1.

\bibitem[\protect\citename{Guo \bgroup et al.\egroup }2014]{guo2014revisiting}
Jiang Guo, Wanxiang Che, Haifeng Wang, and Ting Liu.
\newblock 2014.
\newblock Revisiting embedding features for simple semi-supervised learning.
\newblock In {\em Proceedings of EMNLP}.

\bibitem[\protect\citename{Gurevych}2005]{Gurevych:2005:USC:2145899.2145986}
Iryna Gurevych.
\newblock 2005.
\newblock Using the structure of a conceptual network in computing semantic
  relatedness.
\newblock In {\em Proceedings of IJCNLP}.

\bibitem[\protect\citename{Hassan and
  Mihalcea}2009]{Hassan:2009:CSR:1699648.1699665}
Samer Hassan and Rada Mihalcea.
\newblock 2009.
\newblock Cross-lingual semantic relatedness using encyclopedic knowledge.
\newblock In {\em Proc. of EMNLP}.

\bibitem[\protect\citename{Huang \bgroup et al.\egroup
  }2012]{huang2012improving}
Eric~H Huang, Richard Socher, Christopher~D Manning, and Andrew~Y Ng.
\newblock 2012.
\newblock Improving word representations via global context and multiple word
  prototypes.
\newblock In {\em Proceedings of ACL}.

\bibitem[\protect\citename{Joubarne and
  Inkpen}2011]{Joubarne:2011:CSS:2018192.2018218}
Colette Joubarne and Diana Inkpen.
\newblock 2011.
\newblock Comparison of semantic similarity for different languages using the
  google n-gram corpus and second- order co-occurrence measures.
\newblock In {\em Proceedings of CAAI}.

\bibitem[\protect\citename{Kindermann and Snell}1980]{kindermann80mrf}
Ross Kindermann and J.~L. Snell.
\newblock 1980.
\newblock {\em {Markov Random Fields and Their Applications}}.
\newblock AMS.

\bibitem[\protect\citename{Krahmer \bgroup et al.\egroup
  }2003]{Krahmer:2003:GGR:778822.778825}
Emiel Krahmer, Sebastian van Erk, and Andr{\'e} Verleg.
\newblock 2003.
\newblock Graph-based generation of referring expressions.
\newblock {\em Comput. Linguist.}

\bibitem[\protect\citename{Landauer and Dumais}1997]{landauer:1997}
Thomas~K Landauer and Susan~T. Dumais.
\newblock 1997.
\newblock A solution to plato's problem: The latent semantic analysis theory of
  acquisition, induction, and representation of knowledge.
\newblock {\em Psychological review}.

\bibitem[\protect\citename{Lang and Lapata}2011]{Lang:2011:USR:2145432.2145571}
Joel Lang and Mirella Lapata.
\newblock 2011.
\newblock Unsupervised semantic role induction with graph partitioning.
\newblock In {\em Proceedings of EMNLP}.

\bibitem[\protect\citename{Levy and Goldberg}2014]{levy-goldberg:2014:W14-16}
Omer Levy and Yoav Goldberg.
\newblock 2014.
\newblock Linguistic regularities in sparse and explicit word representations.
\newblock In {\em Proceedings of CoNLL}.

\bibitem[\protect\citename{Mikolov \bgroup et al.\egroup
  }2013a]{mikolov2013efficient}
Tomas Mikolov, Kai Chen, Greg Corrado, and Jeffrey Dean.
\newblock 2013a.
\newblock Efficient estimation of word representations in vector space.
\newblock {\em arXiv preprint arXiv:1301.3781}.

\bibitem[\protect\citename{Mikolov \bgroup et al.\egroup
  }2013b]{mikolov-yih-zweig:2013:NAACL}
Tomas Mikolov, Wen-tau Yih, and Geoffrey Zweig.
\newblock 2013b.
\newblock Linguistic regularities in continuous space word representations.
\newblock In {\em Proceedings of NAACL}.

\bibitem[\protect\citename{Miller}1995]{miller:1995}
George~A Miller.
\newblock 1995.
\newblock Wordnet: a lexical database for english.
\newblock {\em Communications of the ACM}.

\bibitem[\protect\citename{Mnih and Teh}2012]{MnihTeh2012}
Andriy Mnih and Yee~Whye Teh.
\newblock 2012.
\newblock A fast and simple algorithm for training neural probabilistic
  language models.
\newblock In {\em Proceedings of ICML}.

\bibitem[\protect\citename{Myers and Well}1995]{citeulike:8703893}
Jerome~L. Myers and Arnold~D. Well.
\newblock 1995.
\newblock {\em {Research Design \& Statistical Analysis}}.
\newblock Routledge.

\bibitem[\protect\citename{Pennington \bgroup et al.\egroup }2014]{glove:2014}
Jeffrey Pennington, Richard Socher, and Christopher~D. Manning.
\newblock 2014.
\newblock Glove: Global vectors for word representation.
\newblock In {\em Proceedings of EMNLP}.

\bibitem[\protect\citename{Rubenstein and
  Goodenough}1965]{Rubenstein:1965:CCS:365628.365657}
Herbert Rubenstein and John~B. Goodenough.
\newblock 1965.
\newblock Contextual correlates of synonymy.
\newblock {\em Commun. ACM}, 8(10):627--633, October.

\bibitem[\protect\citename{Schuhmacher and Ponzetto}2014]{schuhmacher:2014}
Michael Schuhmacher and Simone~Paolo Ponzetto.
\newblock 2014.
\newblock Knowledge-based graph document modeling.
\newblock In {\em Proceedings of WSDM}.

\bibitem[\protect\citename{Socher \bgroup et al.\egroup
  }2013]{Socher-etal:2013}
Richard Socher, Alex Perelygin, Jean Wu, Jason Chuang, Christopher~D. Manning,
  Andrew~Y. Ng, and Christopher Potts.
\newblock 2013.
\newblock Recursive deep models for semantic compositionality over a sentiment
  treebank.
\newblock In {\em Proceedings of EMNLP}.

\bibitem[\protect\citename{Subramanya \bgroup et al.\egroup
  }2010]{Subramanya:2010:EGS:1870658.1870675}
Amarnag Subramanya, Slav Petrov, and Fernando Pereira.
\newblock 2010.
\newblock Efficient graph-based semi-supervised learning of structured tagging
  models.
\newblock In {\em Proceedings of EMNLP}.

\bibitem[\protect\citename{Talukdar and
  Pereira}2010]{Talukdar:2010:EGS:1858681.1858830}
Partha~Pratim Talukdar and Fernando Pereira.
\newblock 2010.
\newblock Experiments in graph-based semi-supervised learning methods for
  class-instance acquisition.
\newblock In {\em Proceedings of ACL}.

\bibitem[\protect\citename{Turian \bgroup et al.\egroup }2010]{turian:2010}
Joseph Turian, Lev Ratinov, and Yoshua Bengio.
\newblock 2010.
\newblock Word representations: a simple and general method for semi-supervised
  learning.
\newblock In {\em Proc. of ACL}.

\bibitem[\protect\citename{Turney}2006]{Turney:2006:SSR:1174520.1174523}
Peter~D. Turney.
\newblock 2006.
\newblock Similarity of semantic relations.
\newblock {\em Comput. Linguist.}, 32(3):379--416, September.

\bibitem[\protect\citename{Xu \bgroup et al.\egroup }2014]{xu2014rc}
Chang Xu, Yalong Bai, Jiang Bian, Bin Gao, Gang Wang, Xiaoguang Liu, and
  Tie-Yan Liu.
\newblock 2014.
\newblock Rc-net: A general framework for incorporating knowledge into word
  representations.
\newblock In {\em Proceedings of CIKM}.

\bibitem[\protect\citename{Yih \bgroup et al.\egroup }2012]{Yih:2012}
Wen-tau Yih, Geoffrey Zweig, and John~C. Platt.
\newblock 2012.
\newblock Polarity inducing latent semantic analysis.
\newblock In {\em Proceedings of EMNLP}.

\bibitem[\protect\citename{Yu and Dredze}2014]{Yu:2014}
Mo~Yu and Mark Dredze.
\newblock 2014.
\newblock Improving lexical embeddings with semantic knowledge.
\newblock In {\em ACL}.

\bibitem[\protect\citename{Zhu}2005]{Zhu:2005:SLG:1104523}
Xiaojin Zhu.
\newblock 2005.
\newblock {\em Semi-supervised Learning with Graphs}.
\newblock {Ph.D.} thesis, Pittsburgh, PA, USA.
\newblock AAI3179046.

\end{thebibliography}
\bibliographystyle{naaclhlt2015}
\end{document}